\documentclass[letterpaper, 10 pt, conference]{ieeeconf}  % Comment this line out if you need a4paper

\IEEEoverridecommandlockouts                              % This command is only needed if 
\usepackage{times}
\usepackage[utf8]{inputenc} % allow utf-8 input
\usepackage[T1]{fontenc}    % use 8-bit T1 fonts
\usepackage{url}            % simple URL typesetting
\usepackage{booktabs}       % professional-quality tables
\usepackage{amsmath}
\usepackage{amssymb}
\usepackage{amsfonts}       % blackboard math symbols
\usepackage{nicefrac}       % compact symbols for 1/2, etc.
\usepackage{microtype}      % microtypography
\usepackage{subcaption}

\usepackage{color}
\usepackage{graphicx}
\usepackage{adjustbox}
\usepackage[noend]{algpseudocode}
\usepackage{algorithm}
\usepackage{hyperref}       % hyperlinks
\usepackage{xcolor}
\hypersetup{
    colorlinks,
    linkcolor={red!50!black},
    citecolor={blue!50!black},
    urlcolor={blue!80!black}
}
\usepackage{xspace}
\usepackage{wrapfig}
\usepackage{tikz}
\usetikzlibrary{shapes,arrows}

\usepackage{cleveref}

\newtheorem{theorem}{Theorem}

\newtheorem{lemma}{Lemma}

\newtheorem{claim}{Claim}
\newtheorem{example}{Example}

\crefname{section}{Section}{Sections}
\crefname{subsection}{Section}{Sections}
\crefname{definition}{Definition}{Definitions}
\crefname{proposition}{Proposition}{Propositions}
\crefname{lemma}{Lemma}{Lemmas}
\crefname{theorem}{Theorem}{Theorems}
\crefname{corollary}{Corollary}{Corollaries}
\crefname{example}{Example}{Examples}
\crefname{figure}{Figure}{Figures}
\crefname{assumption}{Assumption}{Assumptions}
\crefname{remark}{Remark}{Remarks}
\crefname{running}{Running Example}{Running Examples}
\crefname{algorithm}{Algorithm}{Algorithms}
\crefname{appendix}{Appendix}{appendices}
\crefname{claim}{Claim}{Claims}

\title{Learning Monotone Dynamics by Neural Networks}

\author{Yu Wang, Qitong Gao, and Miroslav Pajic% <-this % stops a space
\thanks{Yu Wang is with the Department of Mechanical and Aerospace Engineering at the University of Florida, Gainesville, FL 32611, USA. Email: {\tt yuwang1@duke.edu}. Qitong Gao and Miroslav Pajic are with the Department of Electrical and Computer Engineering at Duke University, Durham, NC 27708, USA. Emails: {\tt \{qitong.gao, miroslav.pajic\}@duke.edu}}
\thanks{This work is sponsored in part by the ONR under agreements N00014-17-1-2504 and N00014-20-1-2745, AFOSR under award number FA9550-19-1-0169, as well as the NSF CNS-1652544 and CNS-2112562 awards.}% <-this % stops a space
}

\newcommand{\nat}{\mathbb{N}}
\newcommand{\real}{\mathbb{R}}

\renewcommand{\epsilon}{\varepsilon}
\renewcommand{\phi}{\varphi}

\begin{document}

\maketitle

\begin{abstract}
Feed-forward neural networks (FNNs) work as standard building blocks in applying artificial intelligence (AI) to the physical world. They allow learning the dynamics of unknown physical systems (e.g., biological and chemical) {to predict their future behavior}. However, they are likely to violate the physical constraints of those systems without proper treatment. This work focuses on imposing two important physical constraints: monotonicity (i.e., a partial order of system states is preserved over time) and stability (i.e., the system states converge over time) when using FNNs to learn physical dynamics. For monotonicity constraints, we propose to use nonnegative neural networks and batch normalization. For both monotonicity and stability constraints, we propose to learn the system dynamics and corresponding Lyapunov function simultaneously. As demonstrated by case studies, our methods can preserve the stability and monotonicity of FNNs and significantly reduce their prediction errors.
\end{abstract}

\section{Introduction} \label{sec:intro}

Artificial intelligence (AI) is rapidly advancing in the cyber world, especially in computer vision and natural language processing~\cite{goodfellow2016deep}. Recently, there has been a growing interest in building AI that can learn to interact with the physical world~\cite{sunderhauf2018limits}. To this end, feedforward neural networks (FNNs) can serve as building blocks to learn unknown physical system dynamics~\cite{richards_LyapunovNeuralNetwork_2018,chang_NeuralLyapunovControl_2019,kolter_LearningStableDeep_2019} {to predict their future behavior}. Such systems usually obey physical constraints such as monotonicity and stability~\cite{smith_MonotoneDynamicalSystems_2008}.

Monotonicity and stability naturally arise from applications in biology and chemistry. For instance, in an ecological model, the population of several cooperative species can be monotone. If a species' population increases at some time (e.g., by bringing in new ones from outside), then other species' populations will also be higher later, as illustrated in \cref{fig:path of monotone systems}. Besides, monotone systems can also be stable, i.e., the populations converge to given values over time. Examples include traffic networks~\cite{coogan_EfficientFiniteAbstraction_2015}, chemical reactions~\cite{leenheer_MonotoneChemicalReaction_2007}, and bio-ecological models~\cite{costantino_ExperimentallyInducedTransitions_1995}.

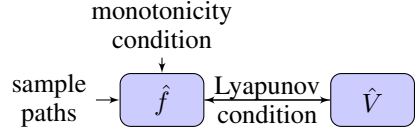
\begin{figure}[t]
\centering
\tikzset{dots/.style args={#1per #2}{line cap=round, dash pattern=on 0 off #2/#1}}
\tikzstyle{block} = [rectangle, draw, fill=blue!20, text width=2.5em, align=center, rounded corners, minimum height=2em]
\tikzstyle{line} = [draw, -latex']
\begin{tikzpicture}
  \node [align=center] (i) {sample\\paths};
  \node [block, right of=i, xshift=0.8cm] (f) {$\hat{f}$};
  \node [align=center, above of=f, yshift=0.1cm] (u) {monotonicity constraint};
  \node [block, right of=f, xshift=2cm] (V) {$\hat{V}$};
  \path [line] (i) -- (f);
  \path [line] (u) -- (f);
  \path [line] (f) -- node [align=center] () {Lyapunov\\condition} (V);
  \path [line] (V) -- (f);
\end{tikzpicture}
\caption{Diagram of learning setup.}
\label{fig:diagram}
\vspace{-20pt}
\end{figure}

However, without proper treatments, monotonicity and stability are likely to be violated by FNNs in learning, and consequently, the learned dynamics will not correctly reflect the dynamics of the real systems. In this work, we propose a new method to impose monotonicity on FNNs in learning without reducing their expressiveness by fusing nonnegative neural networks and batch normalization. In addition, for monotone and stable systems, we propose another method to impose both constraints by simultaneously learning the system dynamics and the corresponding Lyapunov function, as illustrated in \cref{fig:diagram}.

We implement our methods in a window-based fashion on two case studies: the Lotka-Volterra model of two cooperative species that can migrate between multiple patches and the biochemical control circuit of the translation from DNA to mRNA. The results show that our methods ensure the stability and monotonicity of the learned FNNs at most test points. In addition, by imposing the constraints and the window-based implementation, the prediction errors of the FNN to the system are significantly reduced, especially for long time horizons.

\begin{figure}[t]
\centering
\tikzset{dots/.style args={#1per #2}{line cap=round, dash pattern=on 0 off #2/#1}}
\begin{tikzpicture}
  \draw [->, thick] (-0.2, -0.1) -- (-0.2, 1.6);
  \draw [->, thick] (-0.2, -0.1) -- (3.5, -0.1) node[above] () {\small Time};
  \draw [opacity=0] (-2, 0) -- (5, 0);
  \draw [dashed] (0.3, 1.6) node[above] () {$t_0$} -- (0.3, -0.1) ;
  \draw [very thick, dots=4 per 1cm, color=blue] (0.3,1.5) to[out=-45,in=125] (1.2,1) node[above] () {$x'(t)$} to[out=-45,in=160] (2.5,0.3);
  \draw [very thick, dots=4 per 1cm] (0.3,1.3) to[out=-45,in=125] (1.2,0.7) node[below] () {$x(t)$} to[out=-45,in=160] (2.5,0.2);
\end{tikzpicture}
\caption{Monotone system paths. If $x'(t_0) \geq x(t_0)$, then $x'(t) \geq x(t)$ for all $t \geq t_0$.}
\label{fig:path of monotone systems}
\end{figure}

Imposing monotonicity constraints in learning starts from classification and regression of data whose labels increase (or decrease) with the features (e.g., the dependence of the value of a used car on its mileage). To handle such data, monotonicity constraints were imposed for kernel machines~\cite{mukarjee_FeasibleNonparametricEstimation_1994} and trees~\cite{ben-david_MonotonicityMaintenanceInformationTheoretic_1995,neumann_ReliableIntegrationContinuous_2013}. Recently, monotonicity were studied for learning probabilistic dynamical models~\cite{dondelinger2013ode,calderhead2008accelerating,riihimaki2010gaussian,lorenzi2018constraining}. {Our work differs due to the use of the window method and learning with both monotonicity and stability~constraints.}

A common approach is to impose monotonicity as a penalty to the training loss, computed for a set of samples~\cite{gupta_HowIncorporateMonotonicity_2019} 
or the average of some pre-defined distribution~\cite{sill_MonotonicityHints_1997}. {For this approach, the derived NNs are only monotonic for that set of samples or distribution, and require further certification for global monotonicity~\cite{liu2020certified}.} Alternatively, we can indirectly learn a monotone function from its derivative using an NN that only provides nonnegative outputs~\cite{wehenkel_UnconstrainedMonotonicNeural_2019}. However, recovering the monotone function from the trained NN would require integration; thus, any learning errors would accumulate over {the integration/time}, effectively resulting in large approximation errors.

To avoid the above issues, we impose monotonicity through the structure and weights of the NNs {in a correct-by-construction way}. For example, it is proposed to set single weights to be positive~\cite{archer_ApplicationBackPropagation_1993} or introduce constraints between multiple weights~\cite{daniels_MonotonePartiallyMonotone_2010,neumann_ReliableIntegrationContinuous_2013,gupta_MonotonicCalibratedInterpolated_2016}. Examples range from a simple three-layer NN~\cite{sill_MonotonicNetworks_1998} to a more complex structure combining linear calibrators and lattices~\cite{you_DeepLatticeNetworks_2017}. Our work proposes to use two-layer NNs with both min-ReLU and max-ReLU activation functions with nonnegative weights that can capture general nonlinear functions. To avoid sub-optimal outcomes~\cite{marquez2017imposing} caused by the hard constraints in training, we propose to use batch normalization to ``soften'' the constraints. The case studies show that our approach can accurately approximate system dynamics without significantly affecting monotonicity conditions.

% In addition, most of the aforementioned works 
% \cite{archer_ApplicationBackPropagation_1993,sill_MonotonicNetworks_1998,chen_TaylorModelFlowpipe_2012,you_DeepLatticeNetworks_2017,daniels_MonotonePartiallyMonotone_2010} 
% focus on NNs with categorical outputs, while we focus on dynamical systems, where system outputs are~real~vectors. 

Inspired by the idea of using piecewise linear dynamics to approximate (known) nonlinear dynamics in non-learning context~\cite{chen_TaylorModelFlowpipe_2012}, we consider a NN with ReLU activation functions (instead of sigmoid). Similar to~\cite{sill_MonotonicNetworks_1998,daniels_MonotonePartiallyMonotone_2010}, our NN can be viewed as a piecewise linear approximation of a nonlinear function, where the monotonicity is achieved by enforcing positive weights. However, our ReLU NN is more versatile and can have any number of layers, which is needed for learning complex nonlinear dynamics beyond classification and regression~\cite{liang_WhyDeepNeural_2017}.

Our approach to ensuring stability is based on existing work on learning for Lyapunov functions~\cite{richards_LyapunovNeuralNetwork_2018,chang_NeuralLyapunovControl_2019}. Specifically, we simultaneously learn the unknown dynamics and its Lyapunov function. This is similar in spirit to the idea from~\cite{kolter_LearningStableDeep_2019}. However, our update rule for training is different, as the method from~\cite{kolter_LearningStableDeep_2019} does not apply to window-based prediction. Specifically, instead of projecting the dynamics against the learned Lyapunov function~\cite{kolter_LearningStableDeep_2019} to keep them consistent with the Lyapunov condition, we propose to penalize the inconsistency between the learned dynamics and Lyapunov function in training.

\section{Preliminaries} \label{sec:problem}

We consider an unknown discrete-time system
\begin{equation} \label{eq:system}
    x(t + 1) = f(x(t)),
\end{equation}
where $t \in \nat$ is the time, $x \in \mathbb{R}^n$ is the system state, and $f(\cdot)$ is a Lipschitz continuous nonlinear function. For a given initial state, we refer to the corresponding solution $x(t)$ of~\eqref{eq:system} as a \emph{trajectory} of the system. The system~\eqref{eq:system} is a \emph{monotone} system, if it preserves some partial order $\preceq$ on $\real^n$. Here we consider a common partial order~\cite{smith_MonotoneDynamicalSystems_2008}
defined as
\begin{equation} \label{eq:preceq}
    x \preceq y \iff x_i \leq y_i \textrm{ for all } i \in [n],
\end{equation}
where 
$[n] = \{1, \ldots, n\}$,
$x, y \in \real^n$, and
$x_i, y_i \in \real$ denote their $i$-th entry.

The system~\eqref{eq:system} is \emph{monotone} on domain $D \subseteq \real^n$ 
if for any two trajectories $x_1(t), x_2(t) \in D$,
it holds that
\begin{equation}
    x_1(0) \preceq x_2(0) \implies x_1(t) \preceq x_2(t) \textrm{ for all } t \in \nat.
\end{equation}  

The system~\eqref{eq:system} is \emph{monotone}
if and only if the function $f$ is monotonically non-decreasing in the common sense -- i.e., if for any two inputs $x_1, x_2 \in \real^n$,
it holds that $x_1 \preceq x_2 \implies {f} (x_1) \preceq {f} (x_2)$.

\begin{example} \label{ex:scaler monotone system}
A scalar linear system $x(t+1) = a x(t)$ is monotone on the domain $[0, +\infty)$ for any $a \geq 0$, since for any two initial states $x_1 (0) \leq x_2 (0)$, the two corresponding trajectories satisfy $x_1(t) = a^t x_1(0) \leq a^t x_2(0) = x_2(t)$.
Therefore, the ordering between the initial states is preserved during the evolution of two trajectories for all times $t \in \nat$.
\end{example}

It is important to highlight that the monotonicity is defined for the (initial) state not for the time,
as illustrated in \cref{fig:path of monotone systems}. 
By~\cref{ex:scaler monotone system}, when $a \in (0,1)$, the trajectory $x(t) = a^t x(0)$ decreases with the time $t$; yet, the system is still monotone with respect to the initial state $x(0)$. In addition, the monotonicity of the system~\eqref{eq:system} can be equivalently characterized by the gradient of function $f(x)$, as captured in the following lemma from. 

\begin{lemma}{\cite{smith_MonotoneDynamicalSystems_2008}} \label{lem:monotone system}
The system~\eqref{eq:system} is monotone if and only if $\frac{\partial f}{\partial x_i} \geq 0$ for each entry $x_i, (i=1,...,n)$ of $x\in\mathbb{R}^n$. Specially, if the system~\eqref{eq:system} is linear -- i.e., $f(x) = A x$ for some $A \in \real^{n \times n}$, then it is monotone if and only if  each entry $A_{ij}, (i,j\in[n])$ of the matrix $A$ satisfies that $A_{ij}\geq 0$.
\end{lemma}

The system \eqref{eq:system} is \emph{globally asymptotically stable} (or \emph{stable} for short), if it has a (discrete-time) Lyapunov function $V(x)$~\cite{khalil_NonlinearSystems_2002}.
Suppose that $x = 0$ is the stable point of the system (i.e., $f(0) = 0$)  
in general, a stable point $x_0$ can be moved to $0$
by substituting $x$ with $x - x_0$ in the system \eqref{eq:system}. 
Then $V(x)$ should satisfy the \emph{Lyapunov condition} that
\begin{equation} \label{eq:Lyapunov}
    V(0) = 0 \textrm{ and } \forall x \neq 0, V(x) > 0 \textrm{ and } V\left(f(x)\right) - V(x) < 0.
\end{equation}
The Lyapunov function can be viewed as a `potential' with zero value at the stable point and positive values elsewhere. 
For stable systems,
since the \emph{discrete Lie derivative} $V\left(f(x)\right) - V(x)$ is negative,
the (positive) value of the Lyapunov function should decrease along the system path
so that it finally converges to zero at the stable point.   
For monotone stable systems, the Lyapunov function 
can always be written as the maximum of scalar functions~\cite{smith_MonotoneDynamicalSystems_2008}. That is, there exist scalar functions $V_i: \real \to \real$ such~that
\begin{equation} 
\label{eq:monotone Lyapunov}
V(x) = \max_{i \in [n]} V_i (x_i), \quad \textrm{ for } \quad [n] = \{1, \ldots, n\}.
\end{equation}
This provides a foundation for our technique to efficiently learn the Lyapunov function of $\hat{f}$ by FNNs.

\section{Monotone Neural Network}
\label{sec:monotone}

We introduce a window-based method to utilize FNNs to learn the dynamics of the system~\eqref{eq:system}. We show that this method can reduce the learning error in general (\cref{sub:window}), as well as how to impose the monotonicity and stability constraints (\cref{sub:monotonicity constraint,sub:stablity constraint}, respectively). The proofs are available in Appendix of \cite{wang2020deep}.

\subsection{Window-Based Learning Method} 
\label{sub:window}

Conventionally, to learn the dynamics $f$ of the system \eqref{eq:system}, an FNN $\hat{f}_\theta$ parametrized by weights $\theta$ is trained to predict the next state $x(t+1)$ from the current state $x(t)$~\cite{richards_LyapunovNeuralNetwork_2018,chang_NeuralLyapunovControl_2019,kolter_LearningStableDeep_2019}, i.e., 
$$\hat{x}(t+1) = \hat{f}_\theta(x(t)) \approx x(t+1) = f(x(t)).$$ 
To improve the prediction accuracy, we propose a window-based method. Specifically, the FNN $\hat{f}_\theta$ uses a $q$-window of past states to predict the next state, i.e.,
\[
\begin{split}
\hat{x}(t+1) & = \hat{f}_\theta (x(t), x(t-1), \ldots, x(t-q+1)) 
\\ & \approx x(t+1) = f(x(t)).   
\end{split}    
\]
Accordingly, the training loss of $\hat{f}_\theta$ for $f$ is given by
\begin{equation}\label{eq:loss}
    \mathbb{E}_{x(\cdot)\sim \rho}[\Vert x(t+1)-\hat{f}_\theta(x(t),x(t-1),\dots,x(t-q+1)) \Vert ^2],
\end{equation}
where $\rho$ is the state visitation distribution from a random initial state $x(0)$. 
In practice, the expectation in loss~\eqref{eq:loss} is substituted by the empirical training loss for a batch of $N$ given sample paths of the time horizon $H$,~i.e.,
\begin{align} \label{eq:empirical loss}
J(\hat{f}_\theta) = & \ \frac{1}{N} \sum_{i=1}^{N} 
\frac{1}{H - q - 1} \sum_{t = q-1}^{H - 1} 
\big\Vert 
x_i(t+1) - 
\notag \\ & \hat{f}_\theta(x_i(t),x_i(t-1),\dots,x_i(t-q+1)) 
\big\Vert^2,
\end{align}
where $N$ is the batch size and $x_i(\cdot)$ represents the state obtained from the $i^\text{th}$ sample path for all $i\in\{1,...,N\}$.

Admittedly by the dynamics $f$, the next state $x(t+1)$ only depends on the current state $x(t)$. However, using past states can still help training. The past state $x(t-i)$ is related to the next state $x(t+1)$ by $x(t+1) = f^{(i+1)} x(t-i)$. Suppose we start from training $\hat{f}_\theta$ by only using the dependency of $x(t+1)$ on $x(t)$. By adding $x(t-1)$ to training, the FNN $\hat{f}_\theta$ not only needs to fit the dependency of $x(t+1)$ on $x(t)$ but also $x(t+1)$ on $x(t-1)$. This generally reduces the prediction error when $\hat{f}_\theta$ is not exactly equal to $f$. By using the window method, we force the FNN $\hat{f}_\theta$ not only fit with $1$-step dependencies of states by also multi-step dependencies, and hence improves the utility of sample trajectories.

\subsection{Imposing the Monotonicity Constraints} 
\label{sub:monotonicity constraint}

Our method forces the input-output relation of each neuron to be monotone so that the overall FNN is monotone by setting the weights in the FNN to be nonnegative. We achieve this by resetting the negative weights to zero, or to relatively small random numbers that are close to zero, after each backpropagation operation, as in the Dropout~\cite{srivastava2014dropout} method that prevents deep neural networks from overfitting. We refer to such NNs~as~\emph{nonnegative~NNs}.

We use the following rectified linear unit (ReLU) activation functions
\begin{align} 
& \varphi(x_1, \ldots, x_d) = \max \Big\{\sum_{i \in [d]} \theta_i x_i + \theta_0, 0 \Big\} 
\notag 
\\ & \qquad \textrm{or } \min \Big\{\sum_{i \in [d]} \theta_i x_i + \theta_0, 0 \Big\},
\quad \theta_1, \ldots, \theta_d \geq 0, \label{eq:ReLU}
\end{align}
where $x_1, ..., x_d \in \real$ are the inputs to the neurons, $\theta_0$ is the bias, and $\theta_1, \ldots \theta_d$ are the weights of the inputs. This is inspired by the use, in non-learning context, of piecewise linear dynamics to approximate known nonlinear dynamics~\cite{chen_TaylorModelFlowpipe_2012}. Specifically, as $\varphi$ is a piecewise linear function, the FNN $\hat{f}_\theta$ using such activation functions is also piecewise linear. Thus, it can serve as a piecewise linear approximation,  if trained to approximate the dynamics~\eqref{eq:system}. Finally, the min-ReLU activations in~\eqref{eq:ReLU} are needed
to allow for capturing general nonlinear dynamics due to the following claim.

\begin{claim} \label{claim:convex}
If an FNN $\hat{f}_\theta$ only has the max-ReLU activations from \eqref{eq:ReLU},
then $\hat{f}_\theta$ is \emph{convex}.
\end{claim}

Since the activations from \eqref{eq:ReLU} are monotone, the following holds.

\begin{theorem} \label{thm:positive to monotone}
An FNN using the activations from \eqref{eq:ReLU} is monotone.
\end{theorem}

We note that the inverse of \cref{thm:positive to monotone}
may not be necessarily true; i.e., a monotone NN can have negative weights. For example, consider an NN with two hidden layers, each containing a single-neuron. The output of the first hidden layer is $y_1 = \max\{x_1, 1\}$ given its input $x_1$. The ReLU activation in the second hidden layer with a negative weight, computes $z_1 = \max\{- y_1, 1\}$ from the output of the first neuron $y_1$. Yet, the NN is still monotone as the output is always $1$. In addition, the dynamics represented by a monotone NN may decrease over time. For example, consider the single-neuron network that computes $x_1(t+1) = \max\{ 0.5 x_1(t), 1 \}$ from the input $x_1 (t)$. For an initial state $x_1(0) = 10$, the corresponding trajectory
is decreasing with time $t \in \nat$.

\paragraph*{Batch Normalization.} Imposing hard constraints on the weights can lead to undesirable sub-optimal results in training, as observed in~\cite{marquez2017imposing}. Hence, instead of straightly imposing the positive weight constraint $\theta_0, \ldots, \theta_d \geq 0$ for the activations \eqref{eq:ReLU}, we propose to use batch normalization~(BN)~\cite{ioffe2015batch} to soften the constraints and ensure the representation power of $\hat{f}_\theta$. This is because the BN parameters are allowed to converge to optima defined in a broader search space if necessary but can be trained to satisfy the weight constraints as well if it is optimal to do so, although this may lead to tolerable (minor) violations of the hard constraints as we will show in the applications in \cref{sec:case}.

\subsection{Imposing Stability Constraints} 
\label{sub:stablity constraint}

When the system~\eqref{eq:system} of interest is stable, we introduce the following learning method based on an optimization framework~\cite{bezdek2003convergence} that learns $\hat{f}_\theta$ and $\hat{V}_\xi$ iteratively. Recall that the system is stable if and only if it has a Lyapunov function $V(x)$ in the form of \eqref{eq:monotone Lyapunov}. Here, we train an FNN $\hat V_\xi(x)$ of the form~\eqref{eq:monotone Lyapunov} to represent $V(x)$. For a given $\hat{f}_\theta$, we train $\hat V_\xi(x)$ by imposing the Lyapunov condition \eqref{eq:Lyapunov} via the following expected loss
\begin{align} \label{eq:V expected}
& \min_{\xi} 
\mathbb{E}_{x(\cdot) \sim \rho} 
\Big(
\hat{V}_\xi (0)^2 
+ \big[ - \hat{V}_\xi \big( x(t) \big) \big]^+ 
\notag \\ & 
+ \big[ 
\hat{V}_\xi \big( \hat{f}_\theta \big( x(t:t-q+1) \big) \big) - \hat{V}_\xi \big( x(t) \big) 
\big]^+ 
\Big),
\end{align}
where the expectation $\mathbb{E}_{x(\cdot) \sim \rho}$ follows from \eqref{eq:loss}. In \eqref{eq:V expected}, the first term penalizes the non-zero value of $\hat{V}_\xi (0)$, the second term penalizes the negative values of $\hat{V}_\xi (x)$, and the third term penalizes the positive values of the discrete Lie derivative of $\hat{V}_\xi$ for $\hat{f}_\theta$, as discussed in \cref{sec:problem}. Effectively, our approach can be viewed as a discrete-time version of the training loss for the Lyapunov function from~\cite{chang_NeuralLyapunovControl_2019}.

In practice, the expected loss of \eqref{eq:V expected} is approximated by the average of $N$ sample paths of length $H \gg q$ -- i.e., to impose the Lyapunov condition, while training FNN $\hat V_\xi$, we utilize the loss function 
\begin{align} \label{eq:V empirical}
& \min_{\xi} 
\frac{1}{N} \sum_{i=1}^{N} 
\frac{1}{H - q - 1} \sum_{t = q-1}^{H - 1}
\Big(
\hat{V}_\xi (0)^2 
+ \big[ - \hat{V}_\xi \big( x_i(t) \big) \big]^+ 
\notag \\ & 
\qquad + \big[ 
\hat{V}_\xi \big( \hat{f}_\theta \big( x_i(t:t-q+1) \big) \big) - \hat{V}_\xi \big( x_i(t) \big) 
\big]^+ 
\Big).
\end{align}

Similarly, for a given $\hat{V}_\xi$, we train $\hat{f}_\theta$ by incorporating the Lyapunov condition~\eqref{eq:Lyapunov}
into the training loss of \eqref{eq:loss}
\begin{align} \label{eq:f hard}
& \min_{\theta} 
\mathbb{E}_{x(\cdot) \sim \rho} 
\Big( 
\big\Vert f(x(t)) - \hat{f}_\theta(x(t:t-q+1)) \big\Vert^2
\notag \\ & 
\qquad + \Big[
\hat{V}_\xi \big( \hat{f}_\theta \big( x_i(t:t-q+1) \big) \big) 
- \hat{V}_\xi (x) 
\Big]^+
\Big).
\end{align}
% The first 
% Note that in \eqref{eq:f hard}, 
Here,
the first term penalizes the difference in predicting the next state
between $\hat{f}_\theta$ and $f$ from \eqref{eq:system};
the second term penalizes the positive values of the discrete Lie derivative of $\hat{V}_\xi$ for $\hat{f}_\theta$, 
equivalent to the third term of \eqref{eq:V empirical}.
The first two terms of \eqref{eq:V empirical} are not included,
as they are independent~of~$\hat{f}_\theta$.

As done for~\eqref{eq:V empirical}, in practice we approximate the expected loss in \eqref{eq:f hard} 
by using the sample average%  -- i.e., we use
\begin{equation} \label{eq:f hard empirical}
\begin{split}
& \min_{\theta}
\frac{1}{N} \sum_{i=1}^{N} 
\frac{1}{H - q - 1} \sum_{t = q-1}^{H - 1} 
\\ & \quad
\Big( \big\Vert 
f(x_i(t)) - \hat{f}_\theta(x_i(t:t-q+1)) 
\big\Vert^2
\\ & \quad
+ \Big[ 
\hat{V}_\xi \big( \hat{f}_\theta \big( x_i(t:t-q+1) \big) \big) - \hat{V}_\xi \big( x_i(t) \big) 
\Big]^+ 
\Big).
\end{split}
\end{equation}
To train $\hat{f}_\theta$ by \eqref{eq:f hard empirical}, we also impose the monotonicity constraints using the method from Sec.~\ref{sub:monotonicity constraint}.

% It captures the prediction error of $\hat{f}_\theta$,
% whereas the second term penalizes positive values of the Lie derivative.
% \todo{will they know about Lie derivatives? the reviewers might not have taken a nonlinear systems course (like not knowing Lyapunov). if not, introduce here}
%
% \begin{equation} \label{eq:f hard empirical}
%     \min_{\theta \succeq 0} \frac{1}{N} \sum_{i=1}^N 
%     \Big( J(\hat{f}_\theta) + \big[ 
%     \hat{V}_\xi ( \hat{f}_\theta (x) ) - \hat{V}_\xi (x)
%     \big]^+
%     \Big)
% \end{equation}

% Alternatively, \textbf{for \soft method}, 
% we propose a training loss for $\hat{f}_\theta$ by
% \begin{equation} \label{eq:f soft}
% \min_{\theta} 
% \Big(
% J(\hat{f}_\theta) 
% + \Big[ \sup_{x \in \real^n} 
% \Big( \hat{V}_\xi ( \hat{f}_\theta (x) ) - \hat{V}_\xi (x) \Big)
% \Big]^+
% + \max_{j \in [n]} \big[ \sup_{x \in \real^n} - \frac{\partial \hat{f}_\theta}{\partial x_j} (x) \big]^+  
% \Big).
% \end{equation}
% %
% Compared to \eqref{eq:f hard}, the constraint
% on the range of $\theta$ is replaced by the third term 
% that penalizes negative values of the gradients of $\hat{f}_\theta$.
% %
% The corresponding empirical training loss used in practice is
% \begin{equation} \label{eq:f soft empirical}
% \min_{\theta} \frac{1}{N} \sum_{i=1}^N 
% \Big\{ J(\hat{f}_\theta) 
% + \big[ 
% \hat{V}_\xi ( \hat{f}_\theta (x) ) - \hat{V}_\xi (x)
% \big]^+
% + \max_{j \in [n]} \big[ - \frac{\partial \hat{f}_\theta}{\partial x_j} (x_i) \big]^+  
% \Big).
% \end{equation}

% \todo{discuss BN}

\section{Case Studies} \label{sec:case}

To evaluate the effectiveness of our techniques, we consider two high-dimensional complex nonlinear dynamical systems that are monotone due to their physical properties.

\subsection{Lotka-Volterra (LV) Model.} \label{sub:Lotka-Volterra}
We start with the LV model that describes the interaction of 
two cooperative groups (e.g., the males and the females of the same specie)  occupying an environment of $n$ discrete patches.
For the group $k \in \{0, 1\}$, let $x_{ik} (t) \geq 0$ 
be the populations of the group $k$ in the $i^\text{th}$ patch at time $t \in \nat$.
The rate of migration from the $j^\text{th}$ patch to the $i^\text{th}$ patch is $a_{jik} x_{ik} (t)$ 
with $a_{jik} \geq 0$.
At the patch $i$, 
the death rate is $b_{ik} x_{ik} (t)$,
with $b_{ik} \geq 0$;
and the reproduction rate is $c_{ik} x_{ik} (t) x_{i \bar{k}} (t)$,
with $c_{ik} \geq 0$,
where we make the convention that $\bar{k} = (k + 1)\mod 2$.
Thus, for a discrete-time step $\tau > 0$,
the change of the populations of the two groups
at the $i^\text{th}$ patch is given by 
\begin{align} \label{eq:Lotka-Volterra}
& x_{ik} (t + 1) =  x_{ik} (t) +
\tau \Big( c_{ik} x_{ik} (t) x_{i \bar{k}} (t)  
\notag \\ &
- b_{ik} x_{ik} (t) + \sum\nolimits_{ j \in [n] \backslash \{i\} } a_{jik} \big( x_{jk} (t) - x_{ik} (t) \big) \Big).
\end{align}
The LV model is monotone on the positive orthant -- if
the population $x_{ik} (t)$ suddenly increases at time $t$ 
for some $i$ and $k$
(e.g., adding new individuals from the outside),
then the growth rate of all other populations will not decrease;
thus, their populations only increase 
from the increment of $x_{ik} (t)$.~To ensure that the time discretization in~\eqref{eq:Lotka-Volterra} faithfully captures this monotonicity property,
the time step~$\tau$ should satisfy $\tau < 1 / \max_{k \in \{1,2\}} \max_{i \in [n]}  \big( b_{ik} + \sum_{ j \in [n] \backslash \{i\} } a_{jik} \big)$
for the system~\eqref{eq:Lotka-Volterra}~to~be~monotone.

\subsection{Biochemical Control Circuit (BCC) Model.} \label{sub:Biochemical}
We also consider the BCC model describing~the~process of synthesizing a protein from segments of mRNA $E_0$ in a cell, through a chain of enzymes $E_1, ..., E_n$,
where $E_n$ is the end product. 
Let $x_0(t) \geq 0$ be the cellular concentration of mRNA, 
$x_i(t) \geq 0$ be the concentration of the enzyme $i$ for $i \in [n]$ at time $t$.
For each $i \in [n]$, the chemical reaction $\alpha_i E_{i-1} \rightarrow E_i$ 
is assumed to happen with unit rate, where $\alpha_i > 0$.
In addition, the end product stimulates the creation of the mRNA by
the rate $(x_n^p (t) + 1) / (x_n^p (t) + K)$ for some $K > 1$, and $p \in \nat$.
For a discrete-time step $\tau > 0$,
the change of the concentration of the enzyme $i$
is given~by 
\begin{equation} \label{eq:Biochemical}
\begin{split}
& x_0 (t + 1) = x_0 (t) + \tau \Big( \frac{x_n^p (t) + 1}{x_n^p (t) + K} - \alpha_1 x_1 (t) \Big); 
\\ & x_i (t + 1) = x_i (t) + \tau \big( x_{i-1} (t) - \alpha_i x_i (t) \big).
\end{split}
\end{equation}
The BCC model is monotone in the positive orthant since if the concentration of the enzyme $i$ increases with the concentration of the enzyme $i+1$. To ensure that the time discretization in~\eqref{eq:Biochemical} does not violate this monotonicity property, the time step $\tau$ should satisfy $\tau < 1/\max_{i \in [n]} \alpha_i$ for the system~\eqref{eq:Biochemical} to satisfy \cref{lem:monotone system}, i.e., to be monotone.

\begin{table*}[!t]
    \centering
    \caption{Normalized $\ell^2$-norm of Errors in Approximated Trajectories }
    \begin{tabular}{c|c|c|c|c|c|c}
    \hline\hline
     &  \multicolumn{2}{|c|} {\small Monotone \&  Lyapunov}  &  \multicolumn{2}{|c|} {\small Monotone Only} &  \multicolumn{2}{|c} {\small Baseline}\\\hline
     \textbf{\small LV Model}\\\hline
     \small Total Steps\textbackslash Window & \small 100 & \small  1 & \small  100 & \small  1 & \small  100 & \small  1
     \\
    \hline
     \small $1500$    & \small 0.1063 & \small \textbf{0.0514}	& \small\textbf{0.1184} & \small0.5114	& \small\textbf{0.1578} & \small0.5700\\
     \small $2500$    & \small0.1070 & \small\textbf{0.0886}	& \small0.1262 & \small\textbf{0.9383}	& \small\textbf{0.1628} & \small1.0302\\
     \small $3500$    & \small0.1070 & \small\textbf{0.0966} & \small\textbf{0.1616} & \small1.2983 & \small\textbf{0.1970} & \small1.4143\\\hline 
     \textbf{\small BCC Model}\\\hline
     \small Total Steps\textbackslash Window & \small 100 & \small 1 & \small 100 & \small 1 & \small 100 & \small 1
     \\
    \hline
     \small $1500$    & \small\textbf{0.0359} & \small0.2169	& \small\textbf{0.0397} & \small0.2663	& \small\textbf{0.0376} & \small1.6004\\
     \small $2500$    & \small\textbf{0.0334} & \small0.3878	& \small\textbf{0.0856} & \small0.4514	& \small\textbf{0.0349} & \small1.9314\\
     \small $3500$    & \small\textbf{0.0330} & \small0.5543 & \small\textbf{0.1746} & \small0.6290 & \small\textbf{0.0377} & \small2.2409\\
    \hline\hline
    \end{tabular}
    \label{tab:err_norm}
\end{table*}

\subsection{Evaluation.} 
We set $n=10$ in \eqref{eq:Lotka-Volterra} for the LV model and $n=20$ in \eqref{eq:Biochemical} for the BCC model, so the dimensions of the system states are $20$ and $21$, respectively. The training data are drawn from the system with a random set of initial states $x(0)$. More details on the selection of system constants, FNN architectures, and training hyper-parameters are provided in Appendix B of~\cite{wang2020deep}.

We compare the performance when training the FNNs with (I) the proposed loss~\eqref{eq:f hard empirical} that enforces both monotonicity and stability conditions, against (II) monotonicity loss~\eqref{eq:empirical loss} only (which does not ensure stability), and (III) mean-square loss only (i.e., neither monotonicity nor Lyapunov conditions are considered). We test the FNNs by iteratively predicting the system state $T$ time steps (specifically $T=1500,2500,3500$) after a given initial $q$-window of states that are not contained in the training data and then compare with the ground truth. In all cases, the FNN is trained for different windows sizes (specifically $q=1, 100$). The normalized $\ell^2$-norm errors, defined by the ratio of the $\ell^2$-norm of the prediction error to the $\ell^2$-norm of the ground truth, are summarized in Table~\ref{tab:err_norm}. The column headers ``monotone and Lyapunov'', ``monotone only'' and ``baseline'' refer to training with the methods (I), (II), and (III) specified above.

The predicted trajectories for a subset of the states for the two case studies are shown in Figures~\ref{fig:case1_a}(a) and~\ref{fig:case2_a}(a), whereas the results for all states are provided in Appendix D of \cite{wang2020deep}. 
Imposing either the monotonicity or stability constraints reduces the prediction errors, and imposing both brings down the error even further.
In addition, when the stability constraint is imposed (method~(I)), 
the trained FNN becomes much more stable, which significantly 
reduces the prediction errors for long time horizons.
Also, Table~\ref{tab:err_norm} shows using a longer window results in more accurate predictions of future states -- the normalized $\ell^2$-norm of the prediction errors for the window size $q = 100$ is generally much smaller than that for the window size $q = 1$.
In addition, the prediction errors ramp up much slower for      
$q = 100$ than $q = 1$ over long time horizons.

To validate the monotonicity of the trained FNNs, we show in Figures~\ref{fig:case1_a}(b) and~\ref{fig:case2_a}(b) the $x(t+1)$ against $x(t)$ relations for the first $250$ steps for a selection of the dimensions of the states. The figures for all dimensions are given in Appendix D of \cite{wang2020deep}. The monotonicity condition is better satisfied when the monotonicity constraint is imposed in the training loss (in method (I) and (II)), despite the occasional violations due to the batch normalization. Besides, imposing the stability constraint (method (I)) 
generally does not worsen the violation of monotonicity.

\begin{figure*}[t]
\centering
\includegraphics[width=150mm]{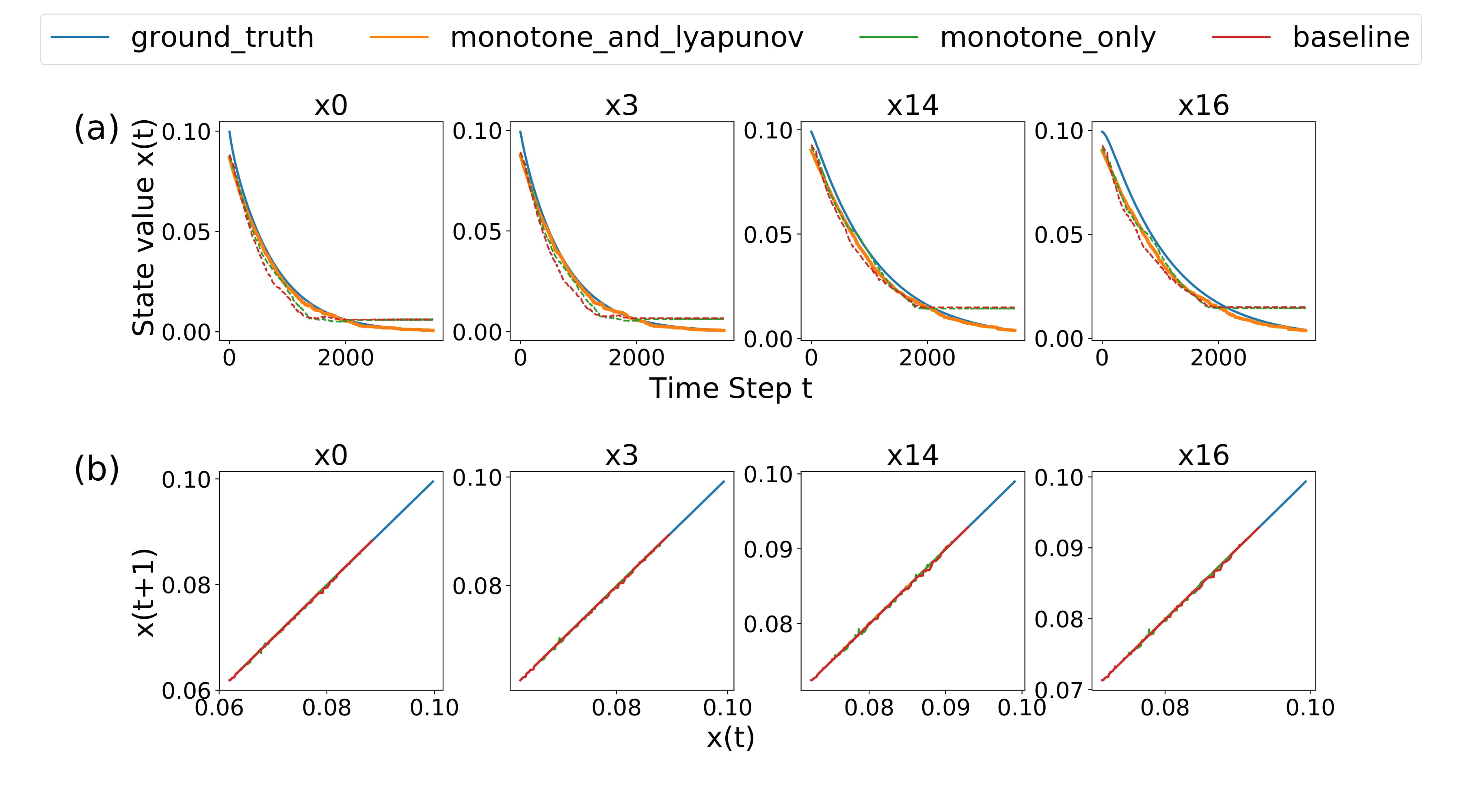}
\vspace{-20pt}
\caption{\small (a) Predicted trajectories of the LV model using $100$-window up to 3500 time steps; (b) The $x(t+1)$-$x(t)$ relation of the predicted LV model trajectory up to 250 time steps.}
% \vspace{-20pt}
\label{fig:case1_a}
\end{figure*}

\begin{figure*}[t]
\centering
\includegraphics[width=150mm]{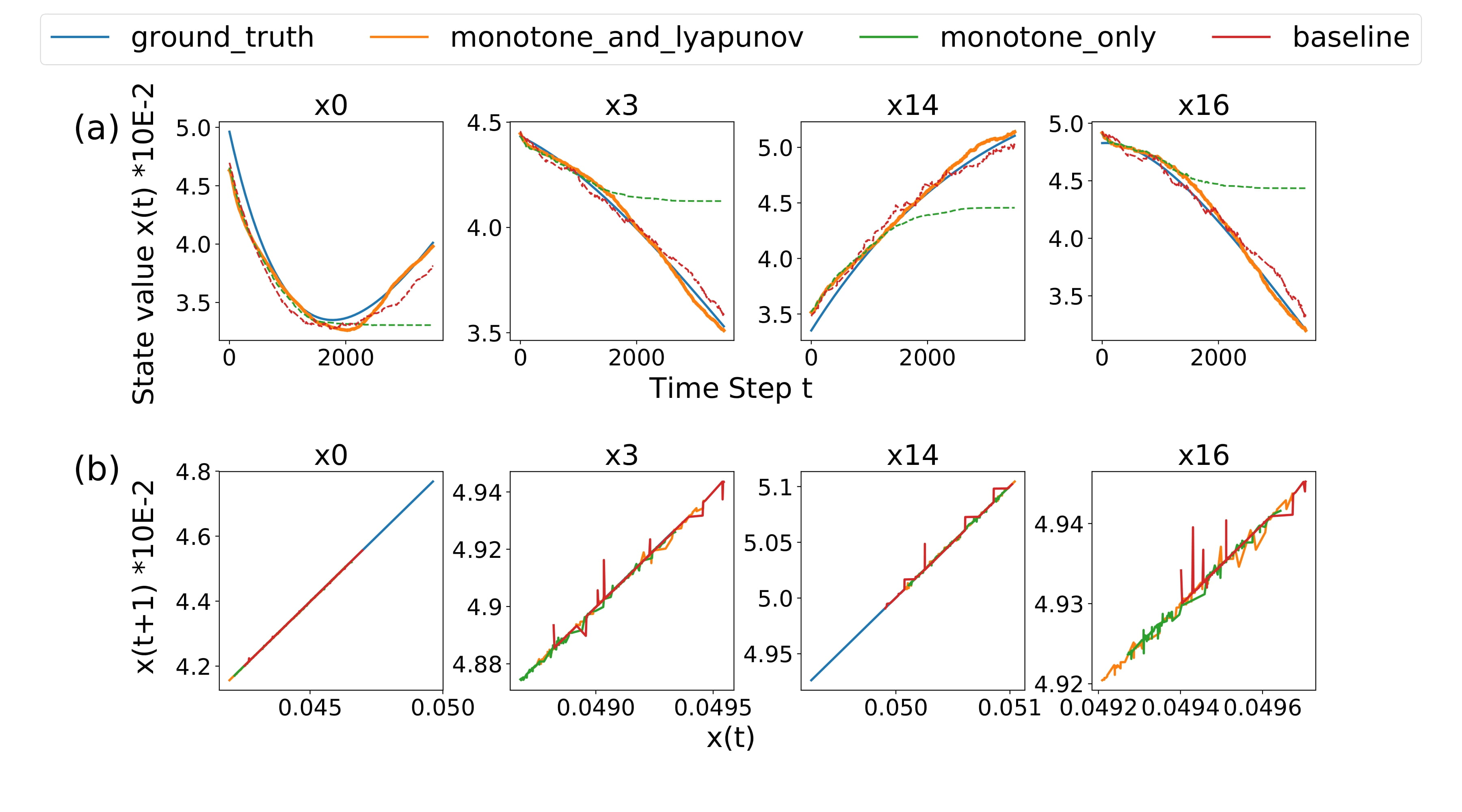}
\vspace{-20pt}
\caption{\small (a) Predicted trajectories of the LV model using $100$-window up to 3500 time steps; (b) The $x(t+1)$-$x(t)$ relation of the predicted LV model trajectory up to 250 time steps.}
% \vspace{-20pt}
\label{fig:case2_a}
\end{figure*}

% \subsection{Field-Noyes Model} \label{sub:Field-Noyes}

% The difference equations below is the discretized Field-Noyes model of 
% the Belousov-Zhabotinski reaction:
% \begin{align} 
%     & x(t + 1) = x(t) + \tau \big( y(t) + x(t) y(t) + x(t) (1 + q x(t)) \big) 
%     \label{eq:Field-Noyes 1}
%     \\ 
%     & y(t + 1) = y(t) + \tau \big( - y(t) + x(t) y(t) + 2 f z(t) \big) 
%     \label{eq:Field-Noyes 2}
%     \\
%     & z(t + 1) = z(t) + \tau \delta \big( x(t) + z(t) \big) 
%     \label{eq:Field-Noyes 3}
% \end{align}
% where $\tau > 0$ is the time step, $f, \delta > 0$, and $q \in (0,1)$ are parameters.
% %
% For $\tau < 1$, the system is monotone in the cone $\{(x, y, z) \in \real^3 \mid x, y, z \geq 0 \}$.

\section{Conclusion} \label{sec:conc}

We introduced a window-based method to learn the dynamics of unknown nonlinear monotone and stable dynamical systems. We employed feedforward neural networks (FNNs) and captured the system's physical properties by imposing the corresponding monotonicity and stability constraints during training. On two high-dimensional complex nonlinear systems (biological and chemical), we showed that the combination of the monotonicity and stability constraints enforces both properties on the learned dynamics while significantly reducing learning errors.

\bibliography{ref}
\bibliographystyle{IEEEtran}

\end{document}